\documentclass[letterpaper, 10 pt, conference]{ieeeconf}  
\IEEEoverridecommandlockouts                              
\usepackage{cite}
\usepackage{graphicx}
\usepackage{float}
\usepackage{hyperref}
\usepackage{amsmath}
\usepackage{amssymb}
\usepackage{booktabs}
\usepackage{multirow}
\usepackage{soul}
\usepackage{url}

\title{\LARGE \bf
DA-GRD: Decision-Aware Grasp-Relevant Disambiguation for tactile recovery under perception-to-execution mismatches
}

\author{Haoran Wang$^{1, 2}$,  Yuteng Sun$^{1, 3}$, Yuanjie Li$^{2}$, Ruofei Bai$^{1,2}$, Meng Yee (Michael) Chuah $^{1}$,\\ Wenyu Liang$^{1}$, jun li$^{1}$, Wei-Yun Yau$^{1,*}$
\thanks{$^{1}$Authors are with the Institute of Advanced Intelligence and Computing (IAIC), Agency for Science, Technology and Research (A*STAR), Singapore.
        }%
\thanks{$^{2}$Nanyang Technological University, Singapore
}
\thanks{$^{3}$Tsinghua University, Beijing, China.
}
\thanks{$^{*}$Correspondence:\nolinkurl{Yau_Wei_Yun@a-star.edu.sg}.
}
}

\begin{document}

\maketitle
\thispagestyle{empty}
\pagestyle{empty}

\begin{abstract}
Grasping is a fundamental robotic capability that bridges perception and physical task execution. This paper studies grasp pose recovery under a perception-to-execution mismatch, where a grasp generated from visual perception may become spatially stale if the object moves before execution, using only sparse tactile interactions and no further visual observations. We propose DA-GRD, Decision-Aware Grasp-Relevant Disambiguation, which maintains a weighted planar belief over possible object configurations and selects tactile probes according to their ability to eliminate hypotheses and improve agreement among candidate task grasps. Rather than fully relocalizing the object, DA-GRD stops when the remaining hypotheses support a common executable grasp. In MuJoCo experiments on ten rigid objects with translations up to 5~cm and yaw perturbations up to $\pm45^\circ$, DA-GRD achieves an 84.7\% physical lift success rate, compared with 9.1\% for stale AnyGrasp, 21.2\% for the original fix-scan baseline, and 63.7\% for fix-scan method adapted with an SE(2) belief. DA-GRD also achieves a 57.3\% Task conditioned Success rate. Across objects, it uses a success-average of 4.13 tactile probes over the ten per-object means, corresponding to a 72.5\% reduction relative to the fixed 15-probe baselines. Real-world experiments on six objects achieve 71.7\% physical lift success and 38.3\% task-conditioned success with 4.20 probes on average. These results show that tactile sensing can recover task-relevant grasps under vision-off conditions with limited physical interaction, without requiring complete object localization.
\end{abstract}

  \begin{figure*}[h]   
      \centering
      \includegraphics[width=\textwidth]{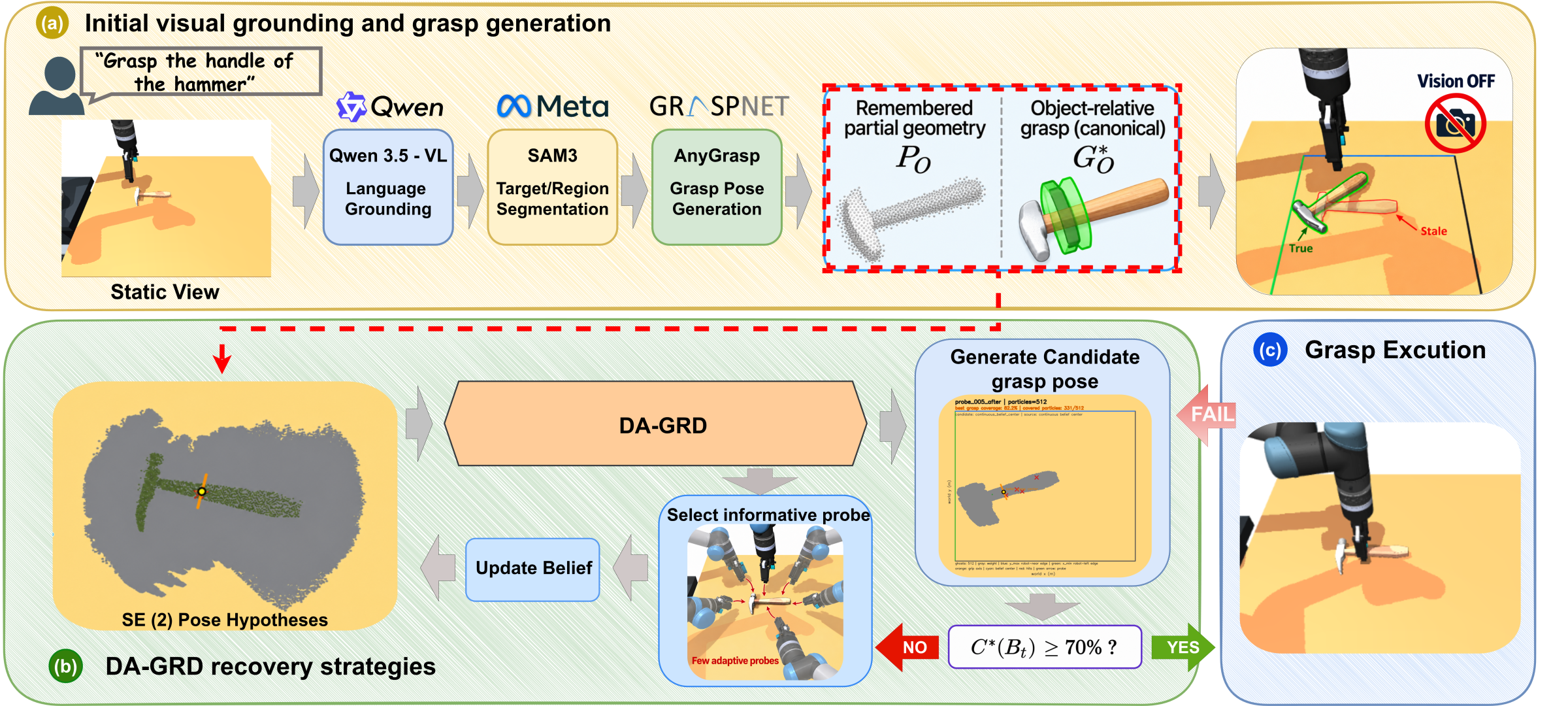}
      \caption{Overview of stale-grasp recovery. (a) A task-conditioned grasp is generated from RGB-D observation and language grounding. (b) After the stale visual grasp fails, its interaction outcome initializes an \(SE(2)\) belief. DA-GRD alternates between evaluating grasp coverage and collecting probe feedback until a sufficiently supported recovery grasp can be executed. (c) Execute the selected grasp pose.}
      \label{fig:overall}
   \end{figure*}
\section{Introduction}

Modern grasping systems can generate 7-DoF grasps directly from RGB-D data~\cite{c1,c2,c3,c4,c5,c6,c7}, while AnyGrasp improves dense grasp generation and temporal robustness~\cite{c9}. Language-conditioned manipulation further allows robots to ground \emph{where} or \emph{what} to grasp from user instructions~\cite{c10,c11}. A practical stack can therefore use language for task semantics and geometric grasp perception for gripper placement.

A grasp may be valid when generated but become incorrect before the grasp is executed, due to perturbations such as the object being moved by a person, prior robot interaction, clutter contact, support motion, or residual registration error, as shown in Fig.\ref{fig:overall}. Closed-loop vision can compensate when the target remains observable~\cite{c8,c9}, but continuous visual feedback is not always available once the hand occludes the scene or enters a confined workspace. In such cases, the semantic intent---e.g., ``grasp the handle''---may still be correct while the stored world-frame pose is stale.
We therefore ask:

\emph{Given a task-conditioned visual grasp that has become spatially stale, can we recover an executable grasp through sparse tactile interactions?}

A natural solution is tactile relocalization. Touch remains effective under occlusion and has long been used for manipulation under pose uncertainty~\cite{c12,c21,c22,c23,c24,c25,c26,c27}. However, every tactile measurement requires motion, time, contact, and may disturb the object. Precise object localization can therefore be unnecessary if the remaining pose hypotheses already imply nearly the same grasp.

This distinction motivates DA-GRD, \emph{Decision-Aware Grasp-Relevant Disambiguation}, an efficient grasp pose recovery mechanism that maintains a weighted planar
belief over possible object configurations and selects tactile
probes according to their ability to eliminate hypotheses and
improve agreement among candidate task grasps.
Specifically, given only the first frame of RGB-D observations, a vision-language module grounds the task-relevant region and AnyGrasp produces an initial grasp $G_W^0$. The object then may undergo an unknown planar displacement with no further vision feedback. 
To actively recover the feasible grasp pose with only tactile interactions, DA-GRD maintains a weighted $SE(2)$ belief over possible current object configurations, which is updated based on the hit, miss, and free-space evidence provided by each tactile probe. Every hypothesis $T_i$ maps the same object-relative task grasp to a candidate world-frame grasp pose $G_i=T_iG_O^*$.

To select the most informative probe candidates, DA-GRD ranks probes using three quantities: posterior mass removable by a miss, newly covered observed surface, and improvement in common grasp support under a counterfactual miss. It stops once a safe grasp covers sufficient posterior mass. Thus, pose uncertainty may remain while action uncertainty is already low.
This design enables reliable grasp pose recovery with significantly fewer tactile probes, thereby improving grasping efficiency when visual observations are limited or prohibited.
Our contributions summarized as follows:
\begin{itemize}
    \item a complete grasp pose generation and recovery framework, featuring initial language-conditioned grasp pose generation and subsequent tactile-based adjustment;
    \item a grasp-relevant $SE(2)$ belief update mechanism that maintains multiple object-pose hypotheses and aggregates them into a committed grasp pose;
    \item DA-GRD, an active tactile policy combining miss-exclusion mass, unexplored observed-surface coverage, and counterfactual grasp-coverage gain for efficient grasp pose recovery.
\end{itemize}

We benchmark the proposed DA-GRD in terms of probing efficiency and task success rate against three baseline recovery strategies: stale AnyGrasp without recovery, the fixed touch-scan particle-filter baseline from \emph{Learning to Grasp Without Seeing}~\cite{c18} (LGWS), and an adapted LGWS baseline with an $SE(2)$ belief representation. Experimental results show that DA-GRD achieves higher success rates on grasp-and-lift tasks while requiring fewer probing interactions, thereby demonstrating its effectiveness in efficient and reliable grasp pose recovery.

\section{Related Work}

\subsection{Visual and Language-Conditioned Grasping}

Grasp perception has progressed from point-cloud grasp detection~\cite{c1}, synthetic grasp learning~\cite{c2}, and large-scale benchmarks~\cite{c3} to direct 6-DoF/7-DoF prediction~\cite{c4,c5,c6,c7}. Closed-loop grasp synthesis can compensate for motion while visual feedback remains available~\cite{c8}, and AnyGrasp provides dense spatially and temporally robust grasp proposals~\cite{c9}. Language-grounded grasping and vision-language-action models further connect semantic instructions to robot actions~\cite{c10,c11}. Our contribution begins after this stage: the visual grasp is meaningful, but becomes spatially stale before execution.

\subsection{Tactile Grasping and Recovery}

Tactile sensing complements vision with local physical evidence. GelSight~\cite{c14} and DIGIT~\cite{c15} support high-resolution contact perception, while tactile feedback has been used for grasp-success prediction and regrasping~\cite{c16,c17}. Murali et al.~\cite{c18} localize an unseen object by sequential planar touch scans and a particle filter before tactile regrasping; we adapt this localization stage as a fixed-scan baseline with the same initial visual prior. Related work combines tactile exploration with grasp refinement~\cite{c19} or uses tactile exploration for unknown-object retrieval in confined spaces~\cite{c20}. Our setting instead starts from a previously selected task-conditioned visual grasp.

\subsection{Active Tactile Exploration}

Active tactile methods choose contact actions to reduce uncertainty. Prior work studies next-best touch~\cite{c21}, uncertainty-aware grasping~\cite{c22}, active Bayesian touch~\cite{c23,c24}, and exploration objectives balancing uncertainty with travel cost~\cite{c25}. For manipulation, Zito et al.~\cite{c26} use tactile information gain for grasp re-planning, Murali et al.~\cite{c27} actively improve visuo-tactile pose registration, and later systems infer object state from tactile sequences~\cite{c28,c29} or learn active exploration for reconstruction~\cite{c30}. These methods primarily measure progress in object space. DA-GRD instead measures whether remaining uncertainty still changes the downstream grasp.

\section{Methodology}

\subsection{Visual Memory and Belief Initialization}

As shown in Fig.~\ref{fig:overall}, DA-GRD begins with an initial
 RGB-D observation and a language instruction specifying the
desired grasp region. A vision-language module grounds the
task-relevant region, and AnyGrasp generates an initial world-frame
grasp $G_W^0$. The corresponding object-relative task grasp $G_O^*$
and remembered partial geometry $P_O$ are stored as visual memory.

After the object undergoes an unobserved planar displacement, visual
feedback is disabled. Rather than immediately starting tactile search,
the robot first executes the original visual grasp $G_W^0$. If this
nominal attempt succeeds, the episode terminates. Otherwise, the
physical outcome of the failed attempt provides the first recovery
observation: contact with the object is recorded as a hit, while an
unsuccessful approach without contact provides miss and free-space
evidence.

DA-GRD uses this initial hit/miss evidence to initialize its weighted
$SE(2)$ belief $\mathcal{B}_0$ over possible current object configurations,
\begin{equation}
    \mathcal{B}_0=\{(T_i,w_i)\}_{i=1}^{N},
\end{equation}
where $T_i$ denotes a possible planar object pose and $w_i$ its weight.
Hypotheses inconsistent with the observation from the nominal
grasp attempt are removed before recovery begins.

At each subsequent recovery step, the current belief is used for two
purposes. First, every pose hypothesis transforms the remembered
task grasp $G_O^*$ into a possible recovery grasp, from which DA-GRD
searches for the grasp with the largest posterior coverage. Second,
the same belief is used to generate and rank probe candidates.
If the best recovery grasp achieves at desirable coverage, the robot executes it. Otherwise, DA-GRD executes the highest-ranked feasible probe, uses the resulting hit or miss to update the belief, and repeats the process.

\begin{figure}[h]
      \centering
      \includegraphics[width=\columnwidth]{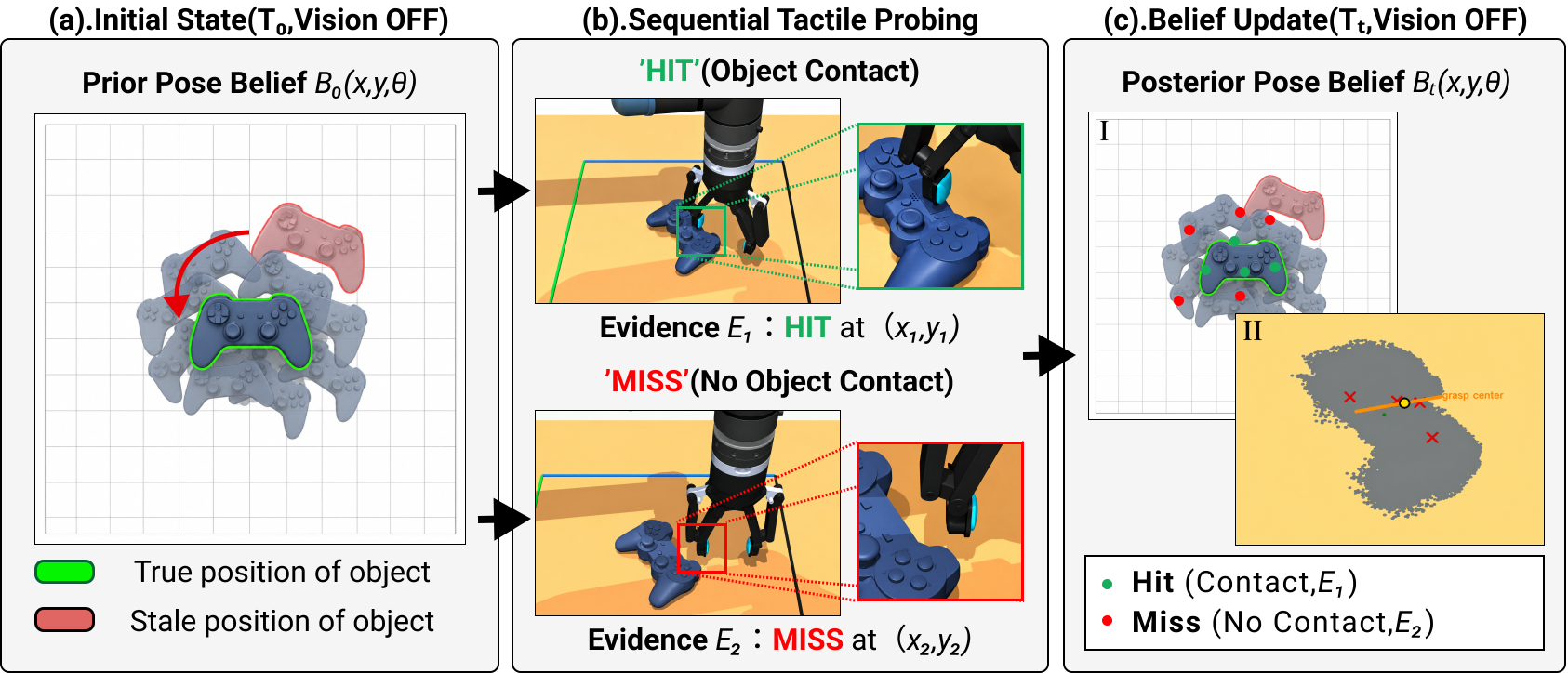}
      \caption{Belief update in DA-GRD. Starting from an initial \(SE(2)\) pose belief, each probe produces either a hit, which supports hypotheses whose surfaces are consistent with the observed contact, or a miss, which removes hypotheses that conflict with the observed free space. The accumulated evidence forms the updated posterior belief.}
      \label{fig:Update}
   \end{figure}
\subsection{Pose-Belief Update}

After executing the selected probe $q_t$, DA-GRD updates the current
belief $\mathcal{B}_t=\{(T_i,w_i)\}$ using the observed hit or miss.

Figure~\ref{fig:Update} illustrates the belief update. For a miss, all hypotheses predicting that the object should have
intersected the swept sensing volume are rejected. The remaining weights are
renormalized to sum to one. The observed free-space region is stored
and remains valid for all later belief updates. For a vertical miss,
only the volume physically swept by the sensing fingers is treated as
free space.

For a hit, the measured contact point $c_t$ is treated as evidence that
some point on the remembered object surface should coincide with the
contact location. Let $p\in P_O$ be a point in the remembered partial
geometry, and let $\theta$ be a candidate hypothesis yaw. DA-GRD searches
for a planar translation $(x,y)$ satisfying
\begin{equation}
    \left\|
    \Pi_{xy}\!\left(T(x,y,\theta)p\right)
    -
    \Pi_{xy}(c_t)
    \right\|
    \leq \epsilon_c ,
    \label{eq:hit_update}
\end{equation}
where $T(x,y,\theta)$ is the candidate planar hypothesis transform,
$\Pi_{xy}(\cdot)$ projects a 3-D point onto the tabletop $xy$ plane,
and $\epsilon_c$ is the contact-matching tolerance. Here, $x$ and $y$ denote the planar translation of the object
reference frame in the world/tabletop coordinate system, and
$\theta$ denotes the object yaw rotation about the world vertical
axis.

Each valid $(x,y,\theta)$ therefore defines a possible object-pose hypothesis
consistent with the new contact. Downward contacts may match any
compatible observed surface, whereas horizontal first contacts must
correspond to a boundary surface of objects.
To account for small probe-induced object motion, DA-GRD also considers
no-motion, sliding, translation, and pivot-like contact interpretations
within $\epsilon_c$. A successor hypothesis is retained only if it remains
consistent with all previous hits, stored free-space constraints,
belief support, and the configured $SE(2)$ support.

\subsection{Probe Candidate Generation}
\begin{figure}[h]
      \centering
      \includegraphics[width=\columnwidth]{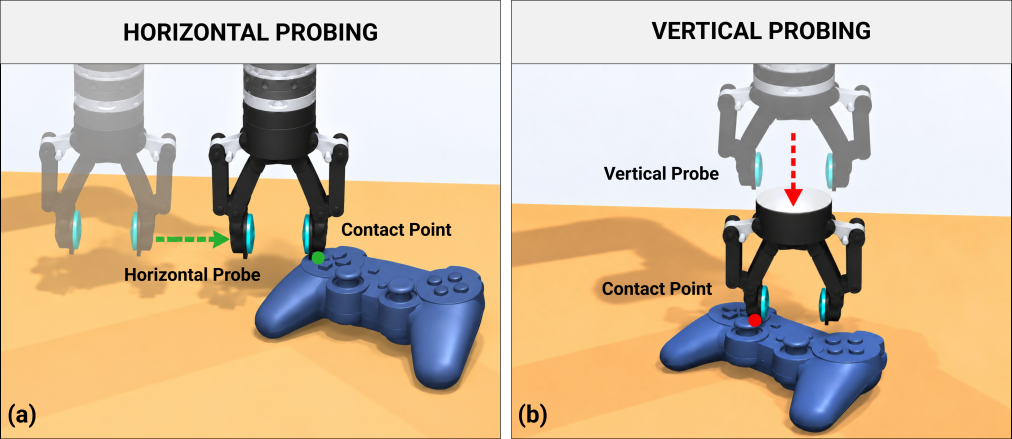}
      \caption{Horizontal and vertical tactile probes used by DA-GRD. (a) Horizontal probes sweep laterally across the belief region and record the first contact. (b) Vertical probes descend from above along the world \(-Z\) direction. Both probe types provide hit/miss and contact evidence for belief updates.}
      \label{fig:probe}
   \end{figure}
After each new hit or miss observation, probe candidates are regenerated
from the updated weighted $SE(2)$ particle belief. For horizontal probing (Fig.~\ref{fig:probe}, left),
we define eight fixed scan directions in the world $xy$ plane, starting
from the world $+X$ axis and spaced every $45^\circ$. For each direction,
the object centers represented by all particles are projected onto the
axis normal to that scan direction. Using the particle weights, three
parallel scan lines are placed at the $0.35$, $0.50$, and $0.65$ weighted
quantiles of the projected distribution, producing $8\times3=24$ basic
horizontal candidates. Their lateral positions therefore adapt to the
current belief, while their start and end points extend along the
corresponding scan direction with a length determined by the spatial
spread of the posterior. Additional spatially distributed routes are
generated from surface regions of the RGB-D partial point
cloud, giving at most 30 horizontal candidates.

Vertical candidates (Fig.~\ref{fig:probe}, right) are generated directly from the current
hypothesis-induced recovery grasps. Each particle transforms the
remembered object-relative task grasp into a possible world-frame grasp,
which is then converted into a vertical probe that moves from its
pre-grasp height to its grasp height along the world $-Z$ direction.
Up to 30 vertical candidates are retained. Thus, after every belief
update, DA-GRD constructs a unified set of at most 30 horizontal and
30 vertical probes for subsequent scoring and selection.

\subsection{Grasp-Relevant Probe Selection}

For a candidate probe $q$, DA-GRD first checks which pose hypotheses
predict that the object would intersect the volume swept by that probe.
The total weight of these hypotheses defines the
\emph{miss-exclusion mass} $M(q)$. Therefore, a large $M(q)$ means
that, if the probe returns a miss, a large portion of the current
$SE(2)$ belief can be rejected.

The second term, \emph{surface novelty} $N(q)$, measures how much
previously unexplored remembered surface is covered by the probe. This
discourages repeatedly probing the same region. Finally, DA-GRD
evaluates whether a miss would make the grasp decision more certain.
After temporarily removing the hypotheses contradicted by such a miss,
the change in the best grasp coverage is defined as
\begin{equation}
    G_{\mathrm{miss}}(q)
    =
    C^*(\mathcal{B}_{q}^{\mathrm{miss}})
    -
    C^*(\mathcal{B}_t),
\end{equation}
where $\mathcal{B}_t$ is the current belief,
$\mathcal{B}_{q}^{\mathrm{miss}}$ is the hypothetical belief after a
miss on $q$, and $C^*(\cdot)$ denotes the best grasp coverage supported
by a belief.

The generated candidates are then ranked by
\begin{equation}
\begin{split}
    \mathrm{Score}(q)=&
    \lambda_M M(q)
    +\lambda_N N(q)
    +\lambda_G G_{\mathrm{miss}}(q)\\
    &-\lambda_t \hat{t}(q)
    -\lambda_r R_{\mathrm{repeat}}(q),
\end{split}
\label{eq:probe_score}
\end{equation}
where $\lambda_M$, $\lambda_N$, and $\lambda_G$ weight the
miss-exclusion mass, surface novelty, and grasp-related gain,
respectively. $\lambda_t$ penalizes estimates motion time $\hat t(q)$, while
$\lambda_r$ penalizes penalizes repeated attempts $R_{\mathrm{repeat}}(q)$. The highest-ranked candidate will be selected to collect new evidence.

Thus, among the generated probe candidates, DA-GRD prefers the one
that can eliminate more plausible object poses, explore new surface
regions, and make the task-conditioned grasp more certain, while
avoiding unnecessarily long or repetitive motions.

\begin{figure}[h]
      \centering
      \includegraphics[width=\columnwidth]{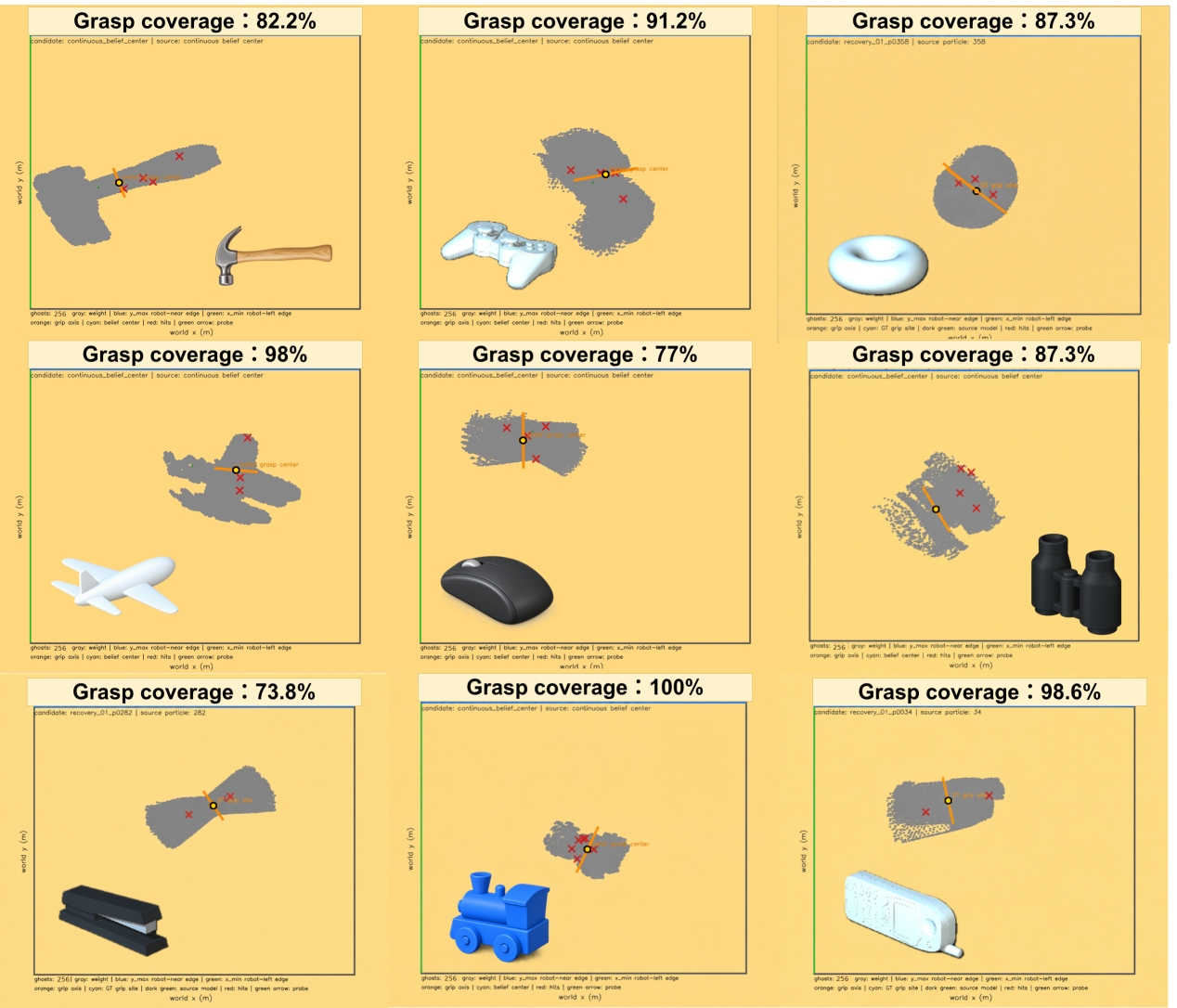}
      \caption{Grasp coverage for the experimental objects. Each subplot displays the current \(SE(2)\) belief and a high-coverage grasp pose. High coverage indicates that, despite remaining object-pose uncertainty, the hypotheses support a common executable grasp.}
      \label{fig:coverage}
   \end{figure}

\begin{table*}[h]
\centering
\caption{Physical lift success rates for DA-GRD, LGWS, and stale AnyGrasp.
Each object is evaluated for 100 episodes.}
\renewcommand{\arraystretch}{1.25}
\setlength{\tabcolsep}{7pt}
\small
\begin{tabular}{c c c c c c c c c c c c c}
\hline
& Airplane & Binoculars & Cube & Game Controller & Hammer & Mouse\\
\hline
AnyGrasp
& 7\% & 0\% & 4\% & 10\% & 7\% & 11\%\\

LGWS
& 16\% & 0\% & 43\% & 10\% & 26\% & 28\%\\

LGWS-A
& 26\% & 18\% & 65\% & 50\% & 84\% & 90\%\\

Ours-PI
& 73\% & 50\% & 89\% & 82\% & 98\% & 86\%\\

Ours-PS
& 68\% & 18\% & 94\% & 82\% & 92\% & 94\%\\
\hline

& Phone &Stapler &Torus & Train &
\multicolumn{2}{c}{\textbf{Overall}} \\
\hline
AnyGrasp
& 13\% & 16\% & 14\% & 9\% &
\multicolumn{2}{c}{\textbf{9.1\%}} \\

LGWS
& 24\%& 22\% & 20\% & 23\% &
\multicolumn{2}{c}{\textbf{21.2\%}} \\

LGWS-A
& 82\%& 94\% & 71\% & 57\% &
\multicolumn{2}{c}{\textbf{63.7\%}} \\

Ours-PI
& 88\%& 97\% & 98\% & 86\% &
\multicolumn{2}{c}{\textbf{84.7\%}} \\

Ours-PS
& 56\% & 96\% & 82\% & 100\%&
\multicolumn{2}{c}{\textbf{78.2\%}}\\
\hline
\end{tabular}
\label{tab:main_success}
\end{table*}

\subsection{Grasp Coverage and Local Refinement}

Each object-pose hypothesis $T_i\in SE(2)$ induces a possible
world-frame realization of the same remembered task grasp:
\begin{equation}
    G_i = T_i G_O^*,
\end{equation}
where $G_O^*$ is the object-relative task grasp and $G_i$ is the grasp
that would be executed if hypothesis $T_i$ were correct.

Although the object poses may still be different, several hypotheses
can support nearly the same grasp. DA-GRD therefore evaluates
uncertainty directly in grasp space. For a candidate grasp $a$, its
posterior coverage is
\begin{equation}
    C(a,\mathcal{B}_t)
    =
    \sum_{i=1}^{N} w_i S(a,G_i),
    \label{eq:grasp_coverage}
\end{equation}
where $w_i$ is the weight of hypothesis $i$, and
$S(a,G_i)=1$ if grasp $a$ is compatible with $G_i$ under the allowed
position, yaw, height, and gripper-width tolerances, and $0$ otherwise.
Thus, $C(a,\mathcal{B}_t)$ measures how much of the current belief
supports grasp $a$.

The best grasp coverage is
\begin{equation}
    C^*(\mathcal{B}_t)
    =
    \max_{a} C(a,\mathcal{B}_t).
    \label{eq:best_grasp_coverage}
\end{equation}
A high $C^*(\mathcal{B}_t)$ means that many remaining object
hypotheses already agree on a common executable grasp, even if the
object pose itself is not fully determined.

Because the $SE(2)$ belief is represented by a finite number of
particles, the induced grasps $\{G_i\}$ are also discrete. The grasp
that is most compatible with the current belief, however, may lie
between these sampled grasps rather than exactly at one of them.

DA-GRD therefore first identifies the largest group of mutually
compatible grasps. If this group contains more than
$\tau_c=0.50$ of the posterior weight, its weighted grasp center is
used only as the starting point for a local search. DA-GRD then
slightly varies the grasp position and yaw around this region and
evaluates each candidate using the same grasp-coverage measure
$C(a,\mathcal{B}_t)$. The current implementation searches with
2-mm translation and approximately $2^\circ$ yaw increments, within
30~mm and $35^\circ$ of the cluster center.

The candidate with the highest posterior coverage is selected. If its
coverage satisfies
\begin{equation}
    C^*(\mathcal{B}_t)\geq \tau_g,
\end{equation}
where $\tau_g$ is the threshold of executable grasp. The robot executes it (as shown in Figure \ref{fig:coverage}); otherwise, tactile probing continues. The cluster center is only used to initialize the local
grasp search. DA-GRD does not average the object poses to estimate a
single object pose. Instead, every refined grasp is evaluated directly
against the current belief.

\section{Experiments and Results}

\subsection{Simulation Setup}

We evaluate in MuJoCo with a UR5 robot arm and a Robotiq 2F-140 gripper. Static RGB-D cameras provide the initial observation; both are disabled after the hidden perturbation. We add contact sensors to the fingertips and outer lateral surfaces to report target-object hit events and contact positions. These sensors do not provide the object's ground-truth pose, evaluation perturbation, contact normal, or hidden mesh state to the planner.

Each episode begins with a language instruction. The visual front end grounds the relevant region and AnyGrasp generates $G_W^0$, which is shared by all methods. The object is then displaced by a hidden planar perturbation. The primary benchmark samples translation within $\pm50$~mm per axis and yaw within the DA-GRD support of $\pm45^\circ$. Regards to the threshold for the grasp coverage, we set $\tau_g=0.70$ to ensure that the executed grasp is supported by a majority of the current belief. The task-conditiond grasp compatibility tolerances are set to 10~mm in position, and $10^\circ$ in yaw. In additon, we set the number of belief as 256 in default. 

\begin{figure}[h]
      \centering
      \includegraphics[width=\columnwidth]{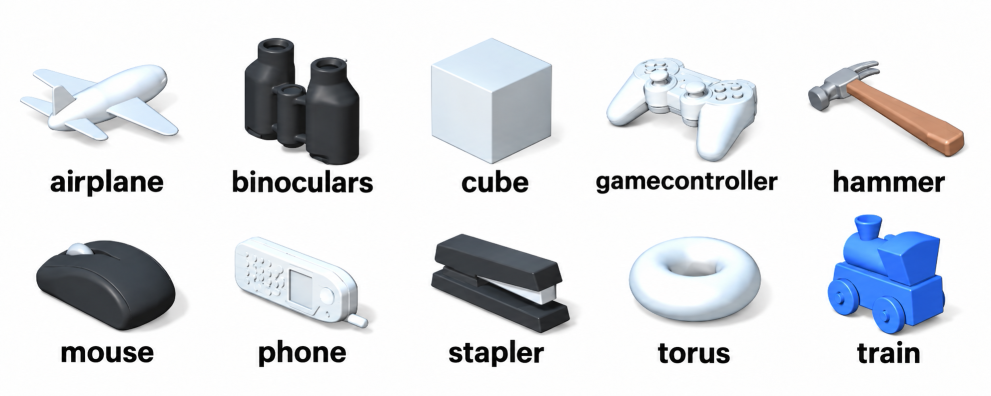}
      \caption{Ten rigid objects used in the evaluation, including elongated, asymmetric, and approximately symmetric geometries.}
      \label{fig:objects}
   \end{figure}

We evaluate all methods under the same in-support full-$SE(2)$ perturbations. The final object set is shown in Fig.~\ref{fig:objects}.

\subsection{Baselines}

\textbf{Stale AnyGrasp. (AG)}
The robot directly executes the initial $G_W^0$ after displacement, with no tactile recovery.

\textbf{LGWS.}
We implement the touch-localization stage of \emph{Learning to Grasp Without Seeing}~\cite{c18} as a fixed-scan XY baseline. The initial visual observation defines a fixed planar workspace, in which the right finger executes 15 uniformly spaced parallel scans at a fixed grasp-plane height, each moving from high to low Y. A scan terminates at first contact or its endpoint. Contact provides occupied-space evidence, while the preceding trajectory, or the full trajectory after a miss, provides free-space evidence. An XY particle filter updates the translation belief after each scan, with a fixed, belief-independent scan order. After all scans, the particle mean is applied to the stale grasp. This baseline estimates translation only and does not use the original learned regrasp network.

\textbf{LGWS adapted. (LGWS-A)}
This variant isolates the effect of representing uncertainty in the
full planar state space. It retains the fixed 15-scan policy of LGWS,
but replaces the original XY-only particle filter with a
256-particle $SE(2)$ belief over
$(\Delta x,\Delta y,\Delta\psi)$. For each scan hit, the measured
first-contact point is incorporated using the same contact-consistency
method used by DA-GRD. The scan order and scan locations remain fixed
and independent of the posterior, and the method does not use
DA-GRD's probe score, grasp-coverage stopping rule, or local grasp
refinement. After all 15 scans, the estimated pose is used to execute
the final grasp.

\textbf{DA-GRD pose-insensitive. (Ours-PI)}
DA-GRD uses the same initial visual memory, contact interface, grasp executor, and success criterion, but maintains an $SE(2)$ belief, actively selects probes with (\ref{eq:probe_score}), and stops by grasp coverage which not sensitive to the precise grasp pose.

\textbf{DA-GRD pose-sensitive. (Ours-PS)}
In the pose-sensitive setting, we retain the same belief update,
probe-selection objective, and local grasp refinement as full DA-GRD,
but replace the grasp-coverage stopping rule with an object-pose
convergence criterion. After each tactile observation, the method
evaluates the weighted translational and circular yaw dispersion of the
current $SE(2)$ particle belief. Exploration terminates only when the
posterior standard deviation is below $4$~mm in translation and
$5^\circ$ in yaw. This criterion uses only the online belief and does
not access the ground-truth object pose.

\subsection{Results and Analysis}

All methods share the same perturbation episodes, initial observations,
task-conditioned grasps, robot controller, and success criterion. We
distinguish three outcomes. \emph{Task-conditioned success} denotes a
successful lift that also satisfies the tolerances around the reference
task-conditioned grasp. \emph{Lift only} denotes a successful lift outside
these tolerances, while \emph{failure} denotes an episode without a
successful lift. Physical lift success is therefore the sum of
task-conditioned success and lift-only episodes.

Table~\ref{tab:main_success} reports physical lift success, where a trial
is successful if the robot grasps the target and lifts it by 5~cm.
Across ten objects, Ours-PI achieves an overall success rate of $84.7\%$,
compared with $9.1\%$ for stale AnyGrasp, $21.2\%$ for the original
LGWS, and $63.7\%$ for LGWS adapted. The pose-sensitive DA-GRD variant
achieves $78.2\%$.

The two LGWS variants isolate the effect of the belief representation.
The original LGWS uses an XY-only particle filter, while LGWS adapted
uses the same 256-particle $SE(2)$ belief as DA-GRD but retains a fixed
15-scan policy. Its improvement from $21.2\%$ to $63.7\%$ shows the
importance of modeling yaw under the present perturbation distribution.
DA-GRD further improves performance by selecting probes according to
their effect on the downstream grasp and stopping when sufficient grasp
coverage is reached. Stale AnyGrasp succeeds mainly when the hidden
motion remains within the gripper capture tolerance, while LGWS benefits
from tactile localization but lacks adaptive probe selection.

\begin{table}[t]
    \centering
    \caption{
        Task-conditioned success rates and tactile interaction counts.
        Each object is evaluated for 100 episodes.}
    \label{tab:gt_success}

    \small
    \renewcommand{\arraystretch}{1.0}

    \begin{tabular*}{\columnwidth}
        {@{\extracolsep{\fill}}lccccc@{}}
        \toprule

        & \multicolumn{5}{c}{Task-Conditioned Success (\%) / Avg. Probes} \\
        \cmidrule(lr){2-6}

        Object & AG & LGWS & LGWS-A & Ours-PI & Ours-PS \\
        \midrule

        Airplane        & 4  & 3  & 13 & 42 / 5.47 & 32 / 6.82 \\
        Bino      & 0  & 0  & 4  & 28 / 7.71 & 14 / 11.08 \\
        Cube            & 2  & 23 & 56 & 58 / 3.34 & 82 / 7.54 \\
        G-C   & 3  & 1  & 30 & 59 / 5.32 & 38 / 8.26 \\
        Hammer          & 5  & 0  & 62 & 74 / 3.00 & 78 / 5.66 \\
        Mouse           & 6  & 11 & 78 & 56 / 4.12 & 80 / 7.44 \\
        Phone           & 7  & 0  & 64 & 65 / 3.06 & 30 / 8.34 \\
        Stapler         & 5  & 0  & 83 & 69 / 3.95 & 82 / 7.62 \\
        Torus           & 12 & 18 & 68 & 72 / 2.28 & 78 / 7.36 \\
        Train           & 5  & 7  & 34 & 50 / 3.07 & 62 / 6.86 \\

        \midrule
        Overall
        & 4.9 & 6.3 & 49.2 & \textbf{57.3 / 4.13}
        & 57.6 / 7.70 \\
        \bottomrule
    \end{tabular*}

    \vspace{1mm}
    \footnotesize
    \raggedright
    Note:  Bino stands for Binoculars  while G-C stands for game controller. AG uses no tactile probes; LGWS and LGWS-A use a fixed budget of
    15 probes. The reported Avg. probes correspond to ours (DA-GRD).
\end{table}

Physical lift success alone does not indicate whether the robot recovers
the task-conditioned grasp specified by the initial visual pipeline.
We therefore also report \emph{task-conditioned success}, which measures
whether the final successful grasp remains consistent with the grasp
generated by the VLM--AnyGrasp pipeline. As shown in
Table~\ref{tab:gt_success}, Ours-PI and Ours-PS achieve overall
task-conditioned success rates of $57.3\%$ and $57.6\%$, respectively,
compared with $4.9\%$ for stale AnyGrasp, $6.3\%$ for the original LGWS,
and $49.2\%$ for LGWS adapted. This shows that DA-GRD can recover not only
a physically successful grasp, but also one consistent with the original
task intent.

DA-GRD also requires substantially fewer tactile interactions. LGWS and
LGWS adapted always execute 15 predefined scans, whereas DA-GRD stops once
the belief sufficiently supports an executable grasp. Across all objects,
Ours-PI requires an average of $4.13$ probes in lift-success episodes,
a $72.5\%$ reduction relative to LGWS. The mean includes both
task-conditioned and lift-only successes, with a successful initial grasp
contributing zero probes. Several objects require even fewer probes,
including torus ($2.28$), hammer ($3.00$), phone ($3.06$), train ($3.07$),
and cube ($3.34$).

Although Ours-PS has a slightly lower overall physical success rate than Ours-PI
(by $6.5\%$), the two achieve similar task-conditioned success. Ours-PS performs
better on several objects with regular geometry and small surface-height
variation, including the hammer, mouse, cube, stapler, and torus. This
highlights the trade-off between the two stopping criteria. Additional
probing can improve pose accuracy and grasp fidelity, but complete pose
convergence is not always necessary. Ours-PI stops once the remaining hypotheses
support a common executable grasp, reducing tactile interaction by
approximately $46.36\%$ compared with Ours-PS. Thus, Ours-PS is more suitable when grasp accuracy is highly sensitive to object
pose, while Ours-PI can reduce interaction and remain high success rate when precise
localization is unnecessary.

The object-wise results also reveal clear failure cases. Both PI and PS
perform relatively poorly on the airplane and binoculars, whose multi-level
and non-planar geometry can produce different 3D contacts with similar
projections onto the tabletop $xy$ plane. Because the current belief is
restricted to planar $SE(2)$ and contact consistency is evaluated after
projection, hypotheses with different contact heights or surface normals may
remain indistinguishable, leading to overestimated grasp coverage or poorly
aligned recovery grasps. A related limitation is the remembered partial point
cloud: missing task-relevant surfaces or similar disconnected regions can
preserve ambiguity even after contact updates. These failure cases motivate
incorporating contact height and surface-normal information, or using a full
3D contact model. Overall, the results support the central motivation of
DA-GRD: complete tactile localization is not always necessary once the
remaining hypotheses sufficiently agree on a task-conditioned grasp.

\begin{figure*}[h]
    \centering
    \includegraphics[width=\textwidth]{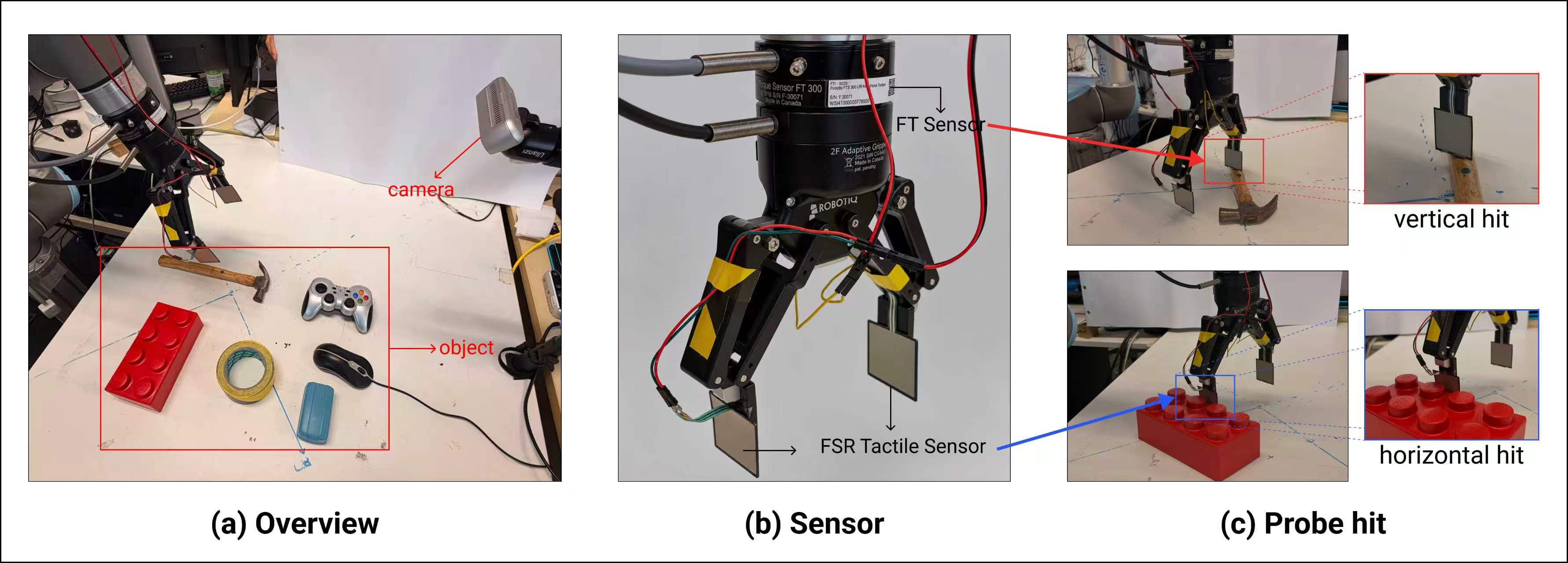}
     \caption{Overall of setup for real world experiment.}
    \label{fig:hardware}
\end{figure*}

\subsection{Real-World Experiment}

We further evaluate DA-GRD on a physical UR5 platform equipped with a
Robotiq 2F-140 gripper. Two FSR sensors are mounted on the gripper
fingers to detect lateral contact during horizontal probing, while a
wrist-mounted FT300 force--torque sensor detects vertical contact. The
visual observation is used to initialize the task-conditioned grasp and
is not updated during tactile recovery.
\begin{figure}[h]
    \centering
    \includegraphics[width=\columnwidth]{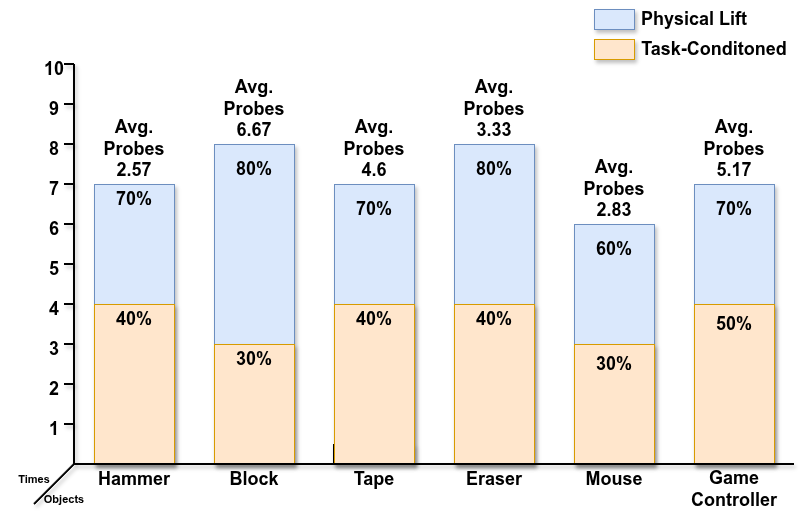}
    \caption{Success rate of real world experiment. Each object is evaluated for 10 episodes. }
    \label{fig:real}
\end{figure}
We evaluate six objects, including a hammer, block, tape, eraser, mouse,
and game controller. Each object is tested for ten episodes. The
physical lift and task-conditioned success rates, together with the
average number of tactile probes, are shown in Fig.~\ref{fig:real}.
DA-GRD achieves physical lift success rates of $70\%$, $80\%$, $70\%$,
$80\%$, $60\%$, and $70\%$ on the hammer, block, tape, eraser, mouse,
and game controller, respectively. The corresponding task-conditioned
success rates are $40\%$, $30\%$, $40\%$, $40\%$, $30\%$, and $50\%$.

The average number of probes varies across objects from $2.57$ to
$6.67$. The hammer requires $2.57$ probes on average, while the block,
tape, eraser, mouse, and game controller require $6.67$, $4.60$, $3.33$,
$2.83$, and $5.17$ probes, respectively. This variation is consistent
with the decision-aware stopping principle: objects whose remaining
pose hypotheses induce similar executable grasps can be recovered with
fewer contacts, whereas objects requiring more precise alignment
continue to receive tactile exploration.

Compared with the simulation results, the real-world success rates are
lower and more variable across objects. Physical contact can cause
unmodeled object motion, while sensor thresholds, calibration errors,
friction, and differences between the remembered geometry and the
physical object affect both belief updates and grasp execution. Despite
these challenges, the real-world results demonstrate that the proposed
belief update, active probe selection, and grasp-aware stopping policy
can be transferred to a physical robot using sparse tactile sensing.

\section{Conclusion}

This work studied stale visual grasp recovery, where a
task-conditioned grasp becomes invalid after an unobserved object
displacement and visual feedback is unavailable. We introduced DA-GRD,
which maintains a weighted $SE(2)$ belief, actively selects tactile
probes, and evaluates uncertainty directly in grasp space. Instead of
requiring full pose convergence, DA-GRD stops once the remaining pose
hypotheses sufficiently support a common executable grasp.

In MuJoCo experiments on ten objects, DA-GRD achieves an overall
physical lift success rate of $84.7\%$ and a task-conditioned success
rate of $57.3\%$, while requiring only $4.13$ tactile probes on average,
a $72.5\%$ reduction relative to the fixed 15-scan LGWS baselines.
Comparison with pose-sensitive further shows the trade-off between
precise localization and interaction efficiency.Experiments on a physical UR5 with six objects further demonstrate that
the belief update, adaptive probing, and grasp-aware stopping strategy
can transfer to real hardware without additional visual feedback.

These results support the main motivation of DA-GRD: complete object
localization is not always necessary when the remaining pose hypotheses
already imply similar grasp actions. The current method is still limited
to planar $SE(2)$ uncertainty and partial remembered geometry, which can
be insufficient for complex non-planar objects. Future work will extend
the framework to 6-DoF motion, richer 3D contact representations, and
learned probe-selection policies for more efficient tactile recovery.

\end{document}